\documentclass[journal]{IEEEtranTIE}
\usepackage{graphicx}
\usepackage{cite}
\usepackage{picinpar}
\usepackage{amsmath}
\usepackage{amsthm}
\usepackage{amssymb}
\usepackage{stfloats}
\usepackage{url}
\usepackage{flushend}
\usepackage[latin1]{inputenc}
\usepackage{colortbl}
\usepackage{soul}
\usepackage{multirow,tabularx}
\usepackage{pifont}
\usepackage{color}
\usepackage{alltt}
\usepackage{caption}
\usepackage{cite}
\usepackage{enumerate}
\usepackage{pbox}
\usepackage{subfig}
\usepackage{bm}
\renewcommand{\figurename}{Fig.}

\newtheorem{al}{Algorithm}

\begin{document}
\title{Canonical Correlation Guided Deep Neural Network}

\author{
	\vskip 1em
	{\color{black}
	Zhiwen Chen, Siwen Mo, Haobin Ke, Steven X. Ding, Zhaohui Jiang, Chunhua Yang, Weihua Gui
	}

	\thanks{
		
		{\color{black}
		
		Zhiwen Chen, Siwen Mo, Zhaohui Jiang, Chunhua Yang and Weihua Gui are with the School of Automation, Central South University, Changsha, 410083, China. Haobin Ke is with the Department of Electrical and Electronic Engineering, the Hong Kong Polytechnic University, Hong Kong. Steven X. Ding are with the Institute for Automatic Control and Complex Systems (AKS), University of Duisburg-Essen, Duisburg, 47057, Germany.

Email addresses of corresponding author: zhiwen.chen@csu.edu.cn.
		}
	}
}
\maketitle

\begin{abstract}
Learning representations of two views of data such that the resulting representations are highly linearly correlated is appealing in machine learning. In this paper, we present a canonical correlation guided learning framework, which allows to be realized by deep neural networks (CCDNN), to learn such a correlated representation. It is also a novel merging of multivariate analysis (MVA) and machine learning, which can be viewed as transforming MVA into end-to-end architectures with the aid of neural networks. Unlike the linear canonical correlation analysis (CCA), kernel CCA and deep CCA, in the proposed method, the optimization formulation is not restricted to maximize correlation, instead we make canonical correlation as a constraint, which preserves the correlated representation learning ability and focuses more on the engineering tasks endowed by optimization formulation, such as reconstruction, classification and prediction. Furthermore, to reduce the redundancy induced by correlation, a redundancy filter is designed. We illustrate the performance of CCDNN on various tasks. In experiments on MNIST dataset, the results show that CCDNN has better reconstruction performance in terms of mean squared error and mean absolute error than DCCA and DCCAE. Also, we present the application of the proposed network to industrial fault diagnosis and remaining useful life cases for the classification and prediction tasks accordingly. The proposed method demonstrates superior performance in both tasks when compared to existing methods. Extension of CCDNN to much more deeper with the aid of residual connection is also presented in appendix.
\end{abstract}

\markboth{None}%
{}

\definecolor{limegreen}{rgb}{0.2, 0.8, 0.2}
\definecolor{forestgreen}{rgb}{0.13, 0.55, 0.13}
\definecolor{greenhtml}{rgb}{0.0, 0.5, 0.0}

\section{Introduction}
\IEEEPARstart{L}{earning} representations of two views of data such that the resulting representations are highly linearly correlated is appealing and a long-term concern in machine learning \cite{Bengio_13,Li_19_multiview}. Canonical correlation analysis (CCA), which was originally proposed for multivariate data analysis \cite{Hotelling,Anderson84}, has attracted much more attention to learn such representations \cite{Hardoon04}. Kernel canonical correlation analysis (KCCA) is an extension of linear CCA in which maximally correlated nonlinear projections, restricted to reproducing kernel Hilbert space with corresponding kernels \cite{scholkopf02}, were developed \cite{akaho2006kernel,Fuku07}. CCA and Kernel CCA have been used for unsupervised data analysis when multiple views are available; learning features for multiple modalities that are then fused for prediction; learning features for a single view when another view is available for representation learning but not at prediction time; and reducing sample complexity of prediction problems using unlabeled data. While kernel CCA allows for learning nonlinear representations, it has the drawback that the representation is limited by the fixed kernel. Also, it is a nonparametric method, the time required to train KCCA or compute the representations of new datapoint scales poorly with the size of the training set. Motivated by these limitations, deep canonical correlation analysis (DCCA) was proposed based on the strong representations learning ability of deep neural networks (DNN) \cite{Andrew13}, which can be viewed as an enhanced version of \cite{Asoh94}. DCCA provides a flexible nonlinear alternative to KCCA, and can learn flexible nonlinear representations as well do not suffer from the aforementioned drawbacks of nonparametric models. Given the empirical success of DNNs on a wide variety of tasks, it is expected to be able to learn more highly correlated representations. In DCCA, two deep nonlinear mappings of two views that are maximally correlated are simultaneously learned. The application of the three methods range broadly across a number of fields, including medicine, meteorology, chemometrics, biology and neurology, natural language processing, speech processing, computer version, multimodal signal processing and industrial fault diagnosis \cite{Anthony92,Sun09,Hardoon04,Juricek2004,Ding2014,Chen_CEP_15,JiangQ_19,ChenH_22,Er2024}. An appealing property of linear CCA for prediction tasks is that, if there is noise in either view that is uncorrelated with other view, the learned representations should not contain the noise in the uncorrelated dimensions \cite{Andrew13}.

CCA, KCCA and DCCA are all using correlation as the optimization formulation \cite{Uurtio_17}. There are different ways to solve this optimization problem by formalizing it as, a standard eigenvalue problem that originally proposed by Hotelling \cite{Hotelling}, a generalised eigenvalue problem \cite{Bach03} and using singular value decomposition (SVD) \cite{Healy1957ARM}. DCCA computes representations of the two views by passing them through multiple stacked layers of nonlinear transformation. The correlation objective is used in DCCA, and it can be optimized using gradient-based optimization. Since the complexity of deep networks, the derivation of the gradient is not entirely straightforward and to be solved with regularization and penalty on parameters.

However, in our recent work of applying DCCA for engineering, we observed that to learn correlation using DNNs is not solely restrict to such a correlation objective. With the strong representation learning of DNNs, the correlation objective can be transformed into a constraint, as shown in Fig. 1, to the conventional objective used in DNNs. For example, reconstruction objective in the unsupervised manner, classification objective and prediction objective in the supervised manner \cite{Li_Lei_2023,Jiang_Ge_2023}. Taking canonical correlation as a constraint of the optimization can also be used for learning correlation. Hence, we name the proposed network as canonical correlation guided deep neural network (CCDNN). The proposed CCDNN has several appealing properties. First, it adds neither extra parameter nor computational complexity. This attributed to that the optimization formulation can be solved using existing solutions. Furthermore, motivated from the work on application of CCA to generate two residual generators (RG) for industrial fault detection \cite{Zhiwen_2017_TIE}, we combine both RGs as a redundancy filter to remove the correlation-induced redundancy in features, which are extracted by the jointly trained DNNs.

Hence, the main features and contributions of this work are:
\begin{itemize}
  \item \textbf{New configuration}. A canonical correlation guided deep neural network is proposed, which is a novel merging of multivariate analysis and machine learning, and focuses more on the specific optimization tasks, like reconstruction, classification and prediction;
  \item \textbf{Redundancy reduction}. The redundancy filter is designed to reduce features' redundancy induced by correlation.
\end{itemize}

In the following sections, we review CCA, KCCA and DCCA, introduce CCDNN, and conduct experiments on three well-known industrial datasets for reconstruction, classification and prediction tasks, respectively. In principle we can evaluate the learned representations on any task in which CCA, KCCA or DCCA have been used. However, in this paper we focus more on the application of the proposed network to industrial fault diagnosis and remaining useful life cases for the classification and prediction tasks accordingly.

The source code and instructions for running the experiments can be accessed at GitHub after the peer-reviewed process. \footnote{Coming soon.}

\section{Background}
\subsection{Linear CCA}
Suppose that the system under consideration has the first view ${\bm{u}}_o\in \mathcal{R}^{l}$, like the input vector, and the second view ${\bm{y}}_o\in\mathcal{R}^{m}$, like the output vector. Assume that
\begin{equation}\label{eq_ccsp}
  \begin{bmatrix}
    {\bm{u}}_o \\
    {\bm{y}}_o \\
  \end{bmatrix}\sim \mathcal{N} \begin{pmatrix}
                                        \begin{bmatrix}
                                        \bm{\mu}_{u}  \\
                                         \bm{\mu}_{y} \\
                                       \end{bmatrix},\begin{bmatrix}
                                                       \mathbf{\Sigma}_{u}& \mathbf{\Sigma}_{uy} \\
                                                       \mathbf{\Sigma}_{uy}^{\text{T}} & \mathbf{\Sigma}_{y} \\
                                                     \end{bmatrix} \\
                                      \end{pmatrix}
\end{equation}
where covariance matrices $\mathbf{\Sigma}_u$ and $\mathbf{\Sigma}_y$ are regular. Denote the mean-centered input and output vectors, respectively, by ${\bm{u}}$ and ${\bm{y}}$, that is
\begin{align}
{\bm{u}}& =({\bm{u}}_{o}-\bm{\mu }_{u})\sim \mathcal{N}(0,\mathbf{\Sigma}_{u})  \label{norm_u} \\
{\bm{y}}& =({\bm{y}}_{o}-\bm{\mu }_{y})\sim \mathcal{N}(0,\mathbf{\Sigma}_{y})  \label{norm_y}
\end{align}
Below, the standard CCA \cite{Anderson84} is introduced. As the basis for the correlation evaluation, matrix
\begin{equation}
\mathbf{\Upsilon}=\mathbf{\Sigma}_{u}^{-1/2}\mathbf{\Sigma }_{uy}\mathbf{%
\Sigma }_{y}^{-1/2}  \label{equ2}
\end{equation}%
is first defined for getting the linear mappings $\mathbf{J}^{\text{T}}\bm{u}$ and $\mathbf{L}^{\text{T}}\bm{y}$ that achieve maximum correlation, i.e. the weight parameters $\mathbf{J}$ and $\mathbf{L}$ are chosen such that the Pearson correlation coefficient $corr(\mathbf{J}^{\text{T}}\bm{u},\mathbf{L}^{\text{T}}\bm{y})$ is maximized. 

Doing a singular value decomposition (SVD), the matrix $\mathbf{\Upsilon}$ can be decomposed as
\begin{equation}
\mathbf{\Upsilon}=\mathbf{\Gamma}\mathbf{\Sigma}\mathbf{R}^{\text{T}}
\label{equ3}
\end{equation}%
with
\begin{equation*}
\mathbf{\Gamma}=(\bm{\gamma}_{1},\ldots ,\bm{\gamma}_{l}),\mathbf{R}=(\bm{\beta}_{1},\ldots , \bm{\beta}_{m}),\mathbf{\Sigma}=%
\begin{bmatrix}
\mathbf{\Sigma}_{\kappa} & 0 \\
0 & 0%
\end{bmatrix}
\end{equation*}%
where $\kappa$ denotes the number of principal components, $\mathbf{\Sigma}_{\kappa}=\text{diag}(\rho_{1},\ldots ,\rho_{\kappa})$, $1\geqslant \rho_{1}\geq \rho_{2}\geq \ldots \geq \rho_{\kappa}\geq0$ are canonical correlation coefficients \cite{Anderson84}. $\rho$ equals $1$ means the highest correlation, $0$ means no correlation. $\bm{\gamma}_{i}$, $i=1,\ldots ,l$ and $\bm{\beta}_{j}$, $j=1,\ldots ,m$ are the corresponding singular vectors.

Therefore, the weight parameters to be solved can be obtained as follows,
\begin{align}
{\mathbf{J}} &= \mathbf{\Sigma}_{u}^{-1/2}\mathbf{\Gamma}\in \mathcal{R}^{l\times l} \label{b51}\\
{\mathbf{L}} &= \mathbf{\Sigma}_{y}^{-1/2}\mathbf{R}\in \mathcal{R}^{m\times m} \label{a51}
\end{align}
which are consist of the canonical correlation vectors. It is well known that the follow important properties of CCA technique hold \cite{Anderson84}
\begin{align}
  {\mathbf{L}}^{\text{T}}\mathbf{\Sigma}_{y}{\mathbf{L}}&=\mathbf{I}_l, \quad {\mathbf{J}}^{\text{T}}\mathbf{\Sigma}_u{\mathbf{J}}=\mathbf{I}_m,\quad
  {\mathbf{J}}^{\text{T}}\mathbf{\Sigma}_{uy}{\mathbf{L}} =\mathbf{\Sigma} \label{eq_jjb} \\
  {\mathbf{L}}^{\text{T}}{\mathbf{\Sigma}}_{uy}^{\text{T}} &= \mathbf{\Sigma}^{\text{T}} {\mathbf{J}}^{\text{T}}{\mathbf{\Sigma}}_{u} \label{CCA_eqa}\\
{\mathbf{J}}^{\text{T}}{\mathbf{\Sigma}}_{uy} &= \mathbf{\Sigma} {\mathbf{L}}^{\text{T}}{\mathbf{\Sigma}}_{y} \label{CCA_eqb}
\end{align}

For easy implementation, an algorithm table is given below,
\begin{al}\label{al_CCA}
CCA algorithm

\noindent \hrulefill

\noindent S1: Center the process data to obtain ${\mathbf{U}}\in\mathcal{R}^{l\times N}$ and ${\mathbf{Y}}\in\mathcal{R}^{m\times N}$, where ${\mathbf{U}}$ and ${\mathbf{Y}}$ are collective data set of ${\bm{u}}_o$ and ${\bm{y}}_o$ with the number of data $N$ ;

\noindent S2: Estimate the covariance and cross-covariance matrices, $\mathbf{\Sigma}_u$, $\mathbf{\Sigma}_y$ and $\mathbf{\Sigma}_{uy}$, respectively \cite{Chen_CEP_15};

\noindent S3: Calculate $\mathbf{J}$, $\mathbf{L}$, $\mathbf{\Sigma}$ using Eqs.(\ref{equ3})-(\ref{a51}).

\noindent \hrulefill

\end{al}

\subsection{KCCA}
Recall that linear CCA looks for linear mappings $\mathbf{J}^{\text{T}}\bm{u}$ and $\mathbf{L}^{\text{T}}\bm{y}$ that achieve maximum correlation. Kernel CCA \cite{akaho2007kernel} extends it by looking for functions $\phi_{1}$ and $\phi_2$ in Hilbert space such that the random variables $\phi_{1}(\bm{u})$ and $\phi_2(\bm{y})$ have maximal correlation. To solve the Kernel CCA problem, the following optimization formulation can be formulated,
\begin{equation}
(\bm{\alpha}_1^{*}, \bm{\alpha}_2^{*})=\mathop{\arg\max}_{(\bm{\alpha}_1,\bm{\alpha}_2)} \quad corr(\phi_1(\mathbf{U}; \bm{\alpha}_1),f_2(\mathbf{Y}; \bm{\alpha}_2))
\end{equation}
where $\bm{\alpha}_1$ and $\bm{\alpha}_2$ are the weight parameters to be obtained by using the kernel trick; $\bm{\alpha}_1^{*}$ and $\bm{\alpha}_2^{*}$ are the optimized weight parameters. Since this solution is now standard, the reader can be referred to \cite{akaho2007kernel} for details.

\subsection{DCCA}
DCCA computes representations of the two views by passing them through multiple stacked layers of nonlinear transformation \cite{Andrew13}. Assume for
simplicity that each intermediate layer in the network for the first view has $c_1$ units, and the final (output) layer has $o$ units. To make notation unconfused, let $\bm{x}_1 \in \mathcal{R}^{n_1}$ be an instance of the first view. Given $\bm{x}_2$ of the second view, the representation $f_2(\bm{x}_2)$ is computed the same way, with different parameters $\mathbf{W}_l^2$ and $b_l^2$. The goal is to jointly learn parameters for both views $\mathbf{W}_l^v$ and $b_l^v$ such that $corr(f_1(\mathbf{X}_1),f_2(\mathbf{X}_2))$ is as high as possible. If $\bm{\theta}_1$ is the vector of all parameters $\mathbf{W}_l^1$ and $b_l^1$ of the first view for $l=1,\cdots,d$, and similarly for $\bm{\theta}_2$, then
\begin{equation}
(\bm{\theta}_1^{*}, \bm{\theta}_2^{*})=\mathop{\arg\max}_{(\bm{\theta}_1, \bm{\theta}_2)} \quad corr(f_1(\mathbf{X}_1;\bm{\theta}_1),f_2(\mathbf{X}_2;\bm{\theta}_2))
\end{equation}

To find the optimized parameters $\bm{\theta}_1^{*}$ and $\bm{\theta}_2^{*}$, we follow the gradient of the correlation objective as estimated on the training data. Let $\mathbf{H}_1 \in \mathcal{R}^{o \times m}$, $\mathbf{H}_2 \in \mathcal{R}^{o \times m}$ be matrices whose columns are representations produced by the deep models on both views, for a training set of size $m$. Let $\bar{\mathbf{H}}_1=\mathbf{H}_1-\frac{1}{m}\mathbf{H}_1\mathbf{1}$ be the centered data matrix (resp. $\bar{\mathbf{H}}_2$), where $\mathbf{1} \in \mathcal{R}^{m\times m}$ is an all-ones matrix, and define $\hat{\mathbf{\Sigma}}_{12}=\frac{1}{m-1}\bar{\mathbf{H}}_1\bar{\mathbf{H}}_2^{\text{T}}$, and $\hat{\mathbf{\Sigma}}_{11}=\frac{1}{m-1}\bar{\mathbf{H}}_1\bar{\mathbf{H}}_1^{\text{T}}+\varepsilon_1I$ for regularization constant $\varepsilon_1$ (resp. $\hat{\mathbf{\Sigma}}_{22}$). Assume that $\varepsilon_1>0$ so that $\hat{\mathbf{\Sigma}}_{11}$ is positive definite.

The total correlation of the top $k$ components of $\mathbf{H}_1$ and $\mathbf{H}_2$ is the sum of the top $k$ singular values of the matrix $\mathbf{R}=\hat{\mathbf{\Sigma}}_{11}^{-1/2}\hat{\mathbf{\Sigma}}_{12}\hat{\mathbf{\Sigma}}_{22}^{-1/2}$. If we take $k=o$, then this is exactly the matrix trace of $\mathbf{R}$, or
\begin{equation}\label{eq_H1}
  corr(\mathbf{H}_1, \mathbf{H}_2)=\|\mathbf{R} \|_{tr}=tr(\mathbf{R}^{\text{T}}\mathbf{R})^{1/2}
\end{equation}

Then, the parameters of DCCA can be trained to optimize this quantity using gradient-based optimization. For detailed information, please refer to \cite{Andrew13}.

\section{Canonical correlation guided deep neural network}
\subsection{Network architecture and its comparison with KCCA and DCCA}
\begin{figure*}[t]
	\begin{center}
		\includegraphics[width=13.4cm]{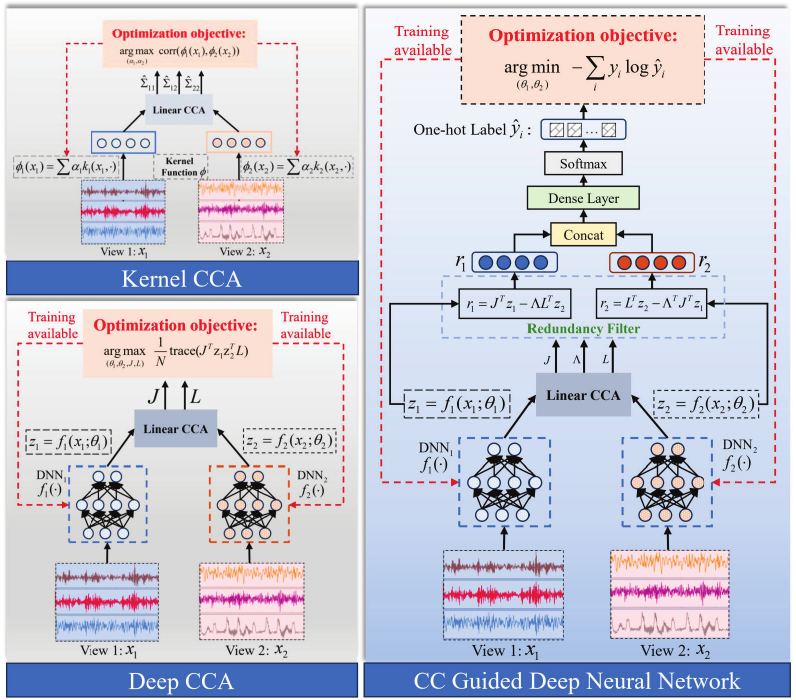}    
		\caption{\rmfamily Architecture comparision of KCCA, DCCA and CCDNN}
		\label{fig_com}
	\end{center}
\end{figure*}

To better understand the architecture of CCDNN, we demonstrate it with the architectures of KCCA and DCCA as shown in Fig. \ref{fig_com}. The difference between KCCA and DCCA are discussed in detail in \cite{Andrew13}. Note that, by comparing with KCCA, CCDNN can also learn flexible nonlinear representations via DNNs, and is a parametric method that training time scales well with data size and the training data need not be referenced when computing the representations of unseen instances. The major difference between DCCA and CCDNN lies in optimization and network architecture. In DCCA, correlation objective is used to guide the training of DNNs. However, learning such a correlation is inherently ill-posed since its solution is not unique. Such a problem is typically mitigated by constraining the solution space by strong prior information \cite{DongC16}. In this study, canonical correlation can be treated as the prior information since our goal is clear that maximizing the correlation between the nonlinear representations via DNNs. Therefore, in our approach, the correlation is transformed into the constraint to restrict the solution space, which enables much more flexible optimization formulation. Hence, three correlations can be learned by changing the optimization formulation in our network, namely reconstruction-oriented correlation that is learned in the unsupervised manner, for instance, with an optimization object of mean square error of the reconstructed views and original views; classification-oriented correlation and regression-oriented correlation that both are learned in a supervised manner. The second difference between DCCA and CCDNN lies in the redundancy filter (RF) to reduce the correlation-induced redundancy in separate mappings of the views. An interesting merit of RF is that it need not to be trained, which means that it doesnot bring computational burden to the training process.

Next we introduce the optimization formulation and the redundancy filter in our approach in detail.

\subsection{Optimization formulation}
Unlike the one in DCCA, taking the reconstruction task as an example, the optimization formulation as shown in Fig. \ref{Task Network} can be defined as follows,
\begin{alignat}{2}
& \mathop{\arg\min}_{(\bm{\theta}_1, \bm{\theta}_2, \bm{\theta}_3, \bm{\theta}_4)}  \nonumber \\ & \frac{1}{N_s} \sum_{k=1}^{N_s}(\|\bm{x}_{1k}-D_1(\mathbf{J}^{\text{T}}f_1(\bm{x}_{1k};\bm{\theta}_1)-\mathbf{\Sigma}\mathbf{L}^{\text{T}}f_2(\bm{x}_{2k};\bm{\theta}_2);\bm{\theta}_3)\|^2 & \nonumber \\
& + \|\bm{x}_{2k}-D_2(\mathbf{L}^{\text{T}}f_2(\bm{x}_{2k};\bm{\theta}_2)-\mathbf{\Sigma}^{\text{T}}\mathbf{J}^{\text{T}}f_1(\bm{x}_{1k};\bm{\theta}_1);\bm{\theta}_4)\|^2) &  \\
\mathrm{s.t.} \quad & [\mathbf{J}, \mathbf{L}, \mathbf{\Sigma}]=CCA(f_1(\mathbf{X}_1;\bm{\theta}_1),f_2(\mathbf{X}_2);\bm{\theta}_2). \nonumber
\end{alignat}
where $\bm{x}_1$ and $\bm{x}_2$ are inputs of view1 and view2, respectively. $\mathbf{X}_1$ and $\mathbf{X}_2$ are collection of $\bm{x}_1$ and $\bm{x}_2$ with $N_s$ training samples, $CCA(\cdot,\cdot)$ denotes the linear canonical correlation analysis operator, which can be achieved using algorithm \ref{al_CCA}.

\begin{figure*}[!t]
	\centering
	\subfloat[DCCA]{
		\includegraphics[width=0.28\linewidth]{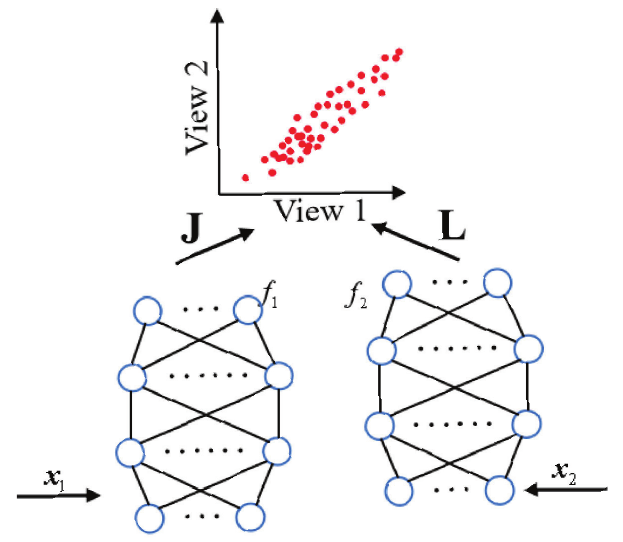}}
	\subfloat[DCCAE]{
		\includegraphics[width=0.30\linewidth]{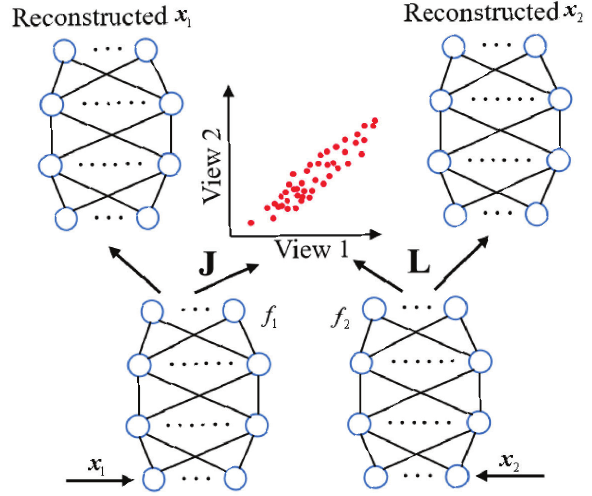}}
    \subfloat[CCDNN]{
		\includegraphics[width=0.25\linewidth]{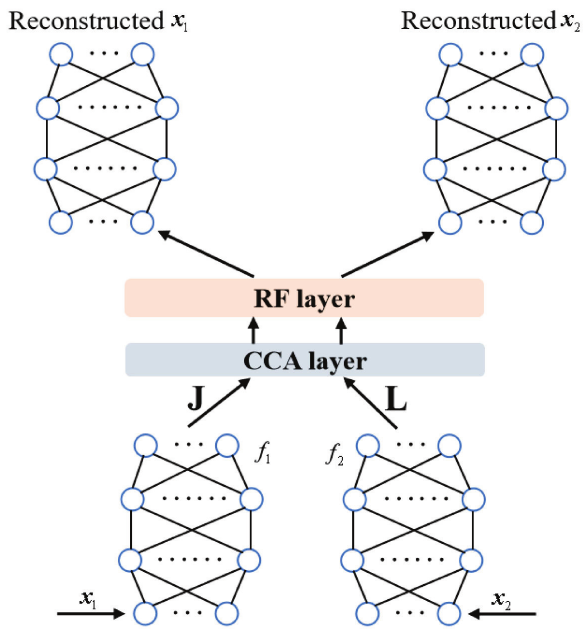}}
	\caption{\rmfamily  Schematic diagram comparison of DCCA, DCCAE and CCDNN}
	\label{three}
\end{figure*}

Taking canonical correlation as a constraint of the loss function add neither extra parameter nor large computational complexity since the trainable parameters in both CCA layer and RF layer are 0s. The entire network can still be trained end-to-end by stochastic gradient degradation (SGD) with back propagation \cite{Lecun98}, and can be easily implemented using common libraries without modifying the solvers. This is not only attractive in practice but also important in the comparisons among DCCA, deep canonically correlated autoencoders (DCCAE) \cite{wang2015deep} and our networks. Fig. \ref{three} illustrates the schematic diagram comparison of them.

The form of the optimization formulation and DNNs are flexible. Experiments in this paper involve optimization formulation as cross-entropy and DNNs as GRU, while other formulation and network architectures are possible. Such a flexible is in general not realized in DCCA.

\subsection{Redundancy filter}
The motivation of redundancy filter is quite straightforward. If the correlation coefficients obtained by the CCA constraint are not all equal to zero, which means that redundancy exists in the outputs (the learned features) of the two deep neural networks due to the correlated relationship. Hence, we design a redundancy filter to reduce such correlation-induced redundancy. It is composed of two parts and can be written as
\begin{align}
    \mathbf{r}_1 &= \mathbf{J}^{\text{T}}\bm{z}_1 - \mathbf{\Sigma}\mathbf{L}^{\text{T}}\bm{z}_2 \label{eq_r1} \\
    \mathbf{r}_2 &= \mathbf{L}^{\text{T}}\bm{z}_2 - \mathbf{\Sigma}^T\mathbf{J}^{\text{T}}\bm{z}_1 \label{eq_r2}
\end{align}
where $\bm{z}_1$ and $\bm{z}_2$ are the outputs of DNN1 and DNN2, respectively. That are $\bm{z}_1 = f_1(\bm{x}_{1};\bm{\theta}_1)$ and $\bm{z}_2 = f_2(\bm{x}_{2};\bm{\theta}_2)$.

Taking the first part in Eq. (\ref{eq_r1}) as an example, it can be rewritten as
\begin{align}
    \mathbf{r}_1 &= \mathbf{J}^{\text{T}}\bm{z}_1 - \mathbf{\Sigma}\mathbf{L}^{\text{T}}\bm{z}_2 \notag \\
    ~ &= \mathbf{\Gamma}^{\text{T}}\mathbf{\Sigma}_{1}^{-1/2}(\bm{z}_1 - \mathbf{\Sigma}_{12}\mathbf{\Sigma}_{2}^{-1}\bm{z}_2) \label{eq_r2s}
\end{align}
where $\mathbf{\Sigma}_{1}$, $\mathbf{\Sigma}_{2}$ and $\mathbf{\Sigma}_{12}$ represent the covariance matrices and cross-covariance of vector $\bm{z}_1$ and $\bm{z}_2$. $\mathbf{\Sigma}_{12}\mathbf{\Sigma}_{2}^{-1}\bm{z}_2$ is a least-mean squares estimation of $\bm{z}_1$ using the data vector $\bm{z}_2$.

To better understand the correlation-induced redundancy, based on properties of CCA, the covariance of the residual signal $\mathbf{r}_1$ can be obtained as
\begin{align}
\mathbf{\Sigma}_{r_1} &= \mathbf{J}^{\text{T}} \text{E}(\bm{z}_1\bm{z}_1^{\text{T}})\mathbf{J}+\mathbf{\Sigma}\mathbf{L}^{\text{T}}\text{E}(\bm{z}_2\bm{z}_2^{\text{T}})\mathbf{L}\mathbf{\Sigma}^{\text{T}}-
\mathbf{J}^{\text{T}}\text{E}(\bm{z}_1\bm{z}_2^{\text{T}})\mathbf{L}\mathbf{\Sigma}^{\text{T}} \notag \\ ~&~ \quad -\mathbf{\Sigma}\mathbf{L}^{\text{T}}\text{E}(\bm{z}_2\bm{z}_1^{\text{T}})\mathbf{J} \notag\\
~ &= \mathbf{I}_{l}-\mathbf{\Sigma}\mathbf{\Sigma}^{\text{T}} \notag\\
~ &= \begin{bmatrix}
       \text{diag}((1-\rho^2_1),\ldots,(1-\rho^2_{\kappa})) & 0 \\
       0 & \mathbf{I}_{l-\kappa} \\
     \end{bmatrix} \in \mathcal{R}^{l\times l}\label{covr1}
\end{align}
It is evident that by using the correlation with $\bm{z}_2$ ($\rho_i\neq 0$), the covariance matrix $\mathbf{\Sigma}_{r_1}$ will decrease.

Analog to $\mathbf{r}_1$, the understanding of $\mathbf{r}_2$ is straightforward, it will not be repeated. From Eqs. (\ref{eq_r1}) and (\ref{eq_r2}), it can be seen that the correlation-induced redundancy is reduced in terms of the covariance of the residual signal, then the filtered signals are concatenated and feed into the dense layer. Note that, if $\text{diag}(\mathbf{\Sigma})=\mathbf{0}$, it is to say that the learned representations of both deep networks has no correlation, which means $\mathbf{r}_1=\mathbf{J}^{\text{T}}\bm{z}_1$ and $\mathbf{r}_2=\mathbf{L}^{\text{T}}\bm{z}_2$, then the redundancy filter passes the inputs without any changes, that is, no redundancy need to be removed.

Note that we have not made any distribution assumption on the two views data other than it being a dataset, this is also apparent difference with linear CCA.

\section{Experiments}

\begin{figure*}[t]
	\begin{center}
		\includegraphics[width=14.4cm]{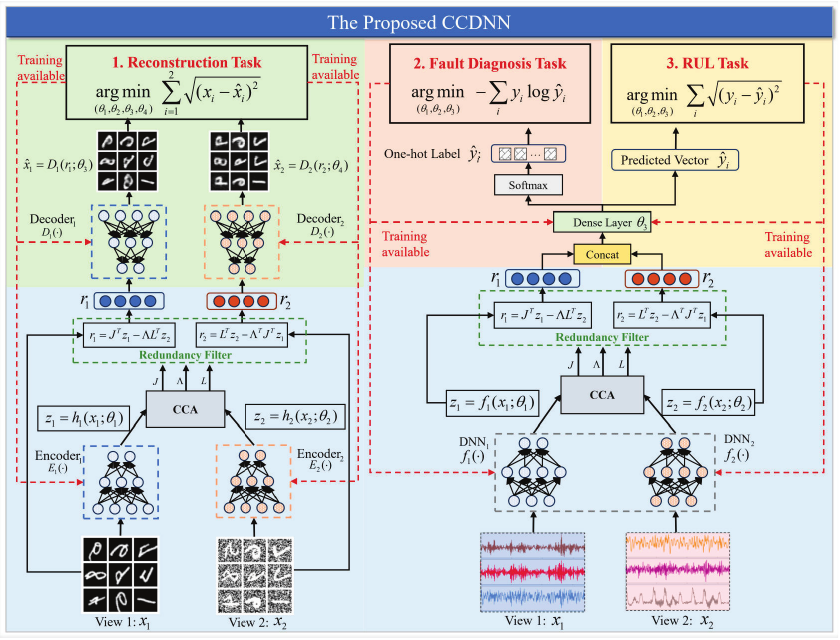}
		\caption{\rmfamily The CCDNN architecture with separate tasks}
		\label{Task Network}
	\end{center}
\end{figure*}

This paper conducts comparative experiments with various methods under the canonical correlation guided learning framework, focusing on several downstream tasks: image denoising reconstruction, industrial fault diagnosis, and remaining useful life prediction (RUL). The network architecture of three downstream tasks is shown in Fig. \ref{Task Network}. Unless otherwise specified, the basic DNN model used by CCDNN is convolutional neural network (CNN), and the hyperparameters for each task are given as follows:

\begin{enumerate}[1)]
	\item Image denoising reconstruction: epochs: 10, batch size: 256, learning rate: 1e-3.
	\item Fault diagnosis: epochs: 100, batch size: 256, learning rate: 1e-3.
	\item RUL: epochs: 100, batch size: total number of samples in the test dataset, learning rate: 1e-3.
\end{enumerate}

\subsection{Reconstruction Task using MNIST dataset}

\begin{figure}[!t]
	\centering
	\subfloat[Normal origin picture]{
		\includegraphics[width=0.39\linewidth]{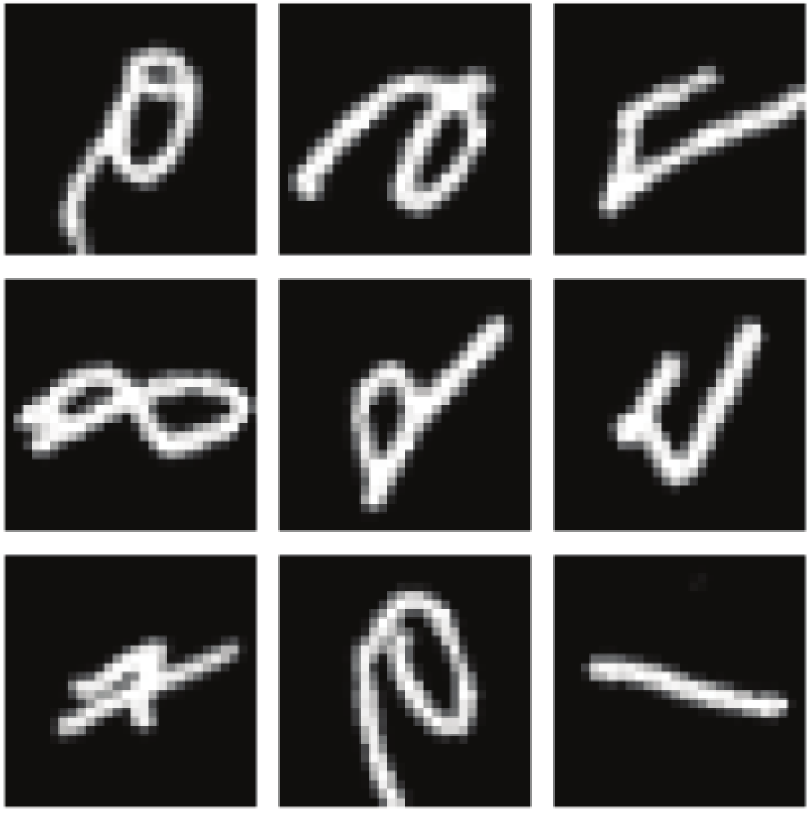}}
	\subfloat[Noisy origin picture]{
		\includegraphics[width=0.40\linewidth]{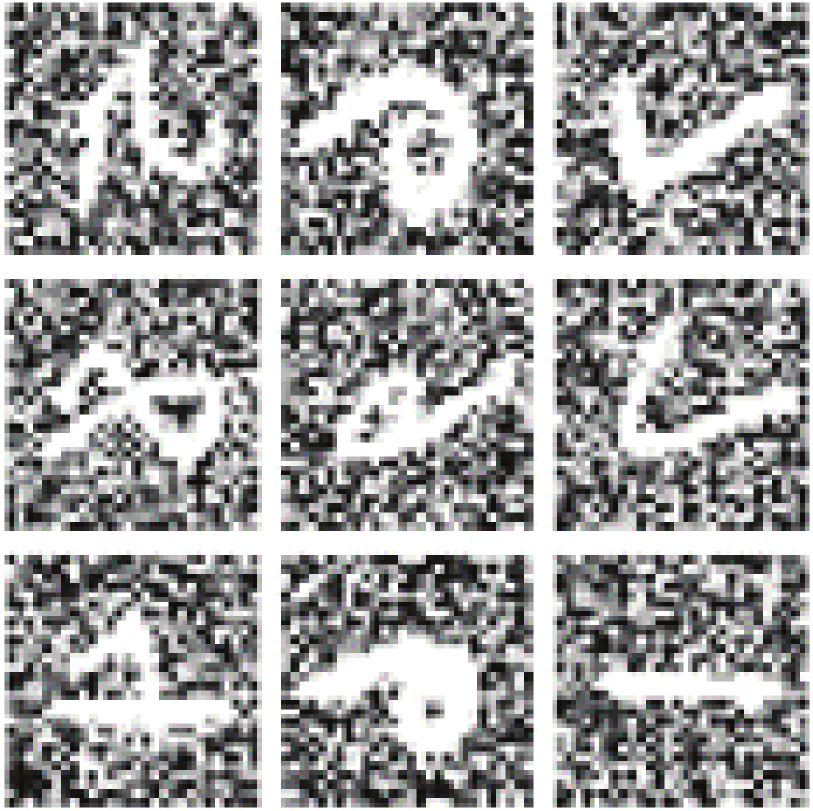}}
	\caption{\rmfamily Image dataset}
	\label{e1}
\end{figure}

In this subsection, the MNIST handwritten digit dataset is used for the image denoising reconstruction task \cite{lecun1998gradient}. This dataset includes 50K training images and 10K testing images. As shown in Fig. \ref{e1}, each image in the original view1 data is a 28*28 grayscale digit image, with pixel values rescaled to [0,1], and the images are randomly rotated by angles in the range of [-${\pi}$/4, ${\pi}$/4] to obtain the view1 input data. Additionally, to construct the Two-view dataset, independent random noise uniformly sampled from [0,1] is added to each pixel of the view1 images, and the pixel values are clipped to [0,1] to obtain the corresponding view2 samples. It can be observed that, due to the interference of noise, observing the view2 images provides no information about the corresponding view1 images given the digit identity. Therefore, a good multi-view learning algorithm should be able to extract features that disregard the noise to achieve denoising and reconstruction of the view2 images \cite{wang2015deep}.

To achieve the above goal, as shown in Fig. \ref{Task Network}, an CCDNN with the deep neural network module as autoencoder is constructed. The autoencoder consists of an encoder and a decoder, both of which are CNN. First, each view1 image and the corresponding view2 image are separately input into the encoder network to obtain low-dimensional embeddings $\bm{z}_1$ and $\bm{z}_2$ mapped in the latent space (also known as the encoding space); then $\bm{z}_1$ and $\bm{z}_2$ are input into the redundancy filter to get two residual signals $\bm{r}_1$ and $\bm{r}_2$; the decoder maps the residual signals back to the original data space, resulting in the reconstructed image data $\bm{{\hat x}}_1$ and $\bm{\hat x}_2$. During the training process, parameters of the encoder and decoder are optimized by minimizing the loss function as described in Eq. (14).

To verify the superiority of the proposed method in image reconstruction capability, two classical deep CCA frameworks, DCCA and DCCAE, as well as CCDNN without the redundancy filter (${\text{CCDNN}_{\text{wRF}}}$), were compared. The metrics for evaluating the reconstruction capability of different methods are mean squared error (MSE) and mean absolute error (MAE), defined as follows:

\begin{align}
	MSE = \frac{{\sum\limits_{i = 1}^n {{{|| {{x_i} - {{\hat x}_i}} ||}^2}} }}{n}  \label{e12}
\end{align}

\begin{align}
	MAE = \frac{{\sum\limits_{i = 1}^n {\left| {{x_i} - {{\hat x}_i}} \right|} }}{n}
	\label{e13}
\end{align}
where ${x_i}$ is the true value of the $i$th sample, ${\hat x_i}$ is the model's prediction for the $i$th sample, and $n$ is the total number of samples. The smaller the values of MSE and MAE, the better the effectiveness of model reconstruction.

The metric for measuring the correlation between two features extracted by the model is the total correlation, which is defined as follows \cite{chen2021comparative}:

\begin{align}
	\mathop {\max }\limits_{{{\bf{\Theta }}_h},{{\bf{\Theta }}_g} } \frac{1}{N}{\rm{trace}}({\mathbf{J}^{\text{T}}}{\bf{U}}{{\bf{Y}}^{\text{T}}}\mathbf{L})
	\label{e13}
\end{align}

\begin{figure}[t]
	\centering
	\subfloat[DCCA Normal Reconstruction]{
		\includegraphics[width=0.40\linewidth]{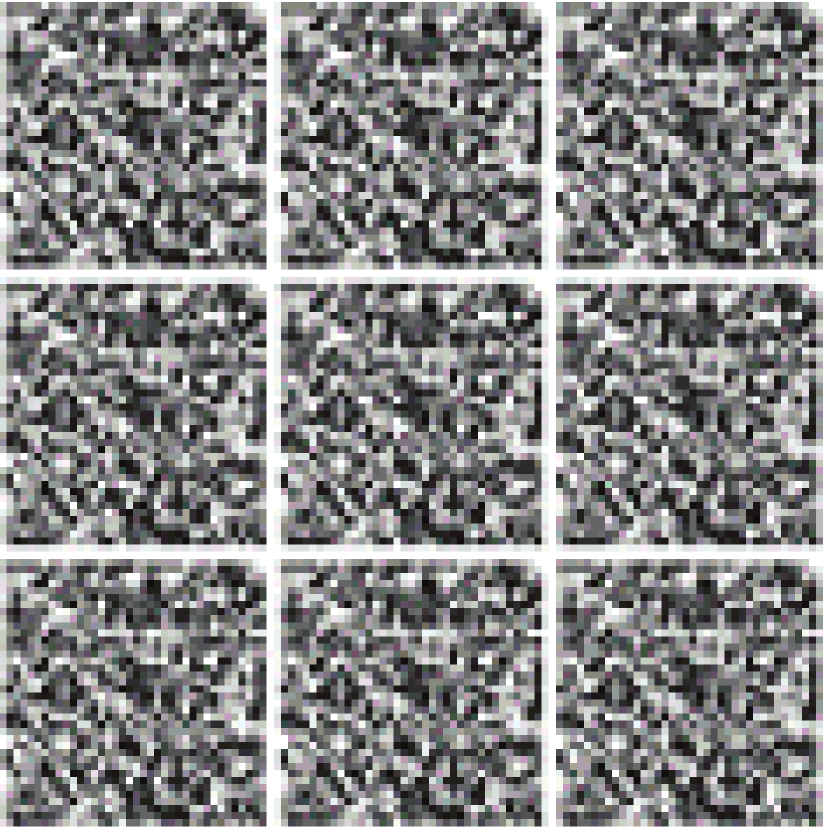}}
	\subfloat[DCCA Noisy Reconstruction]{
		\includegraphics[width=0.40\linewidth]{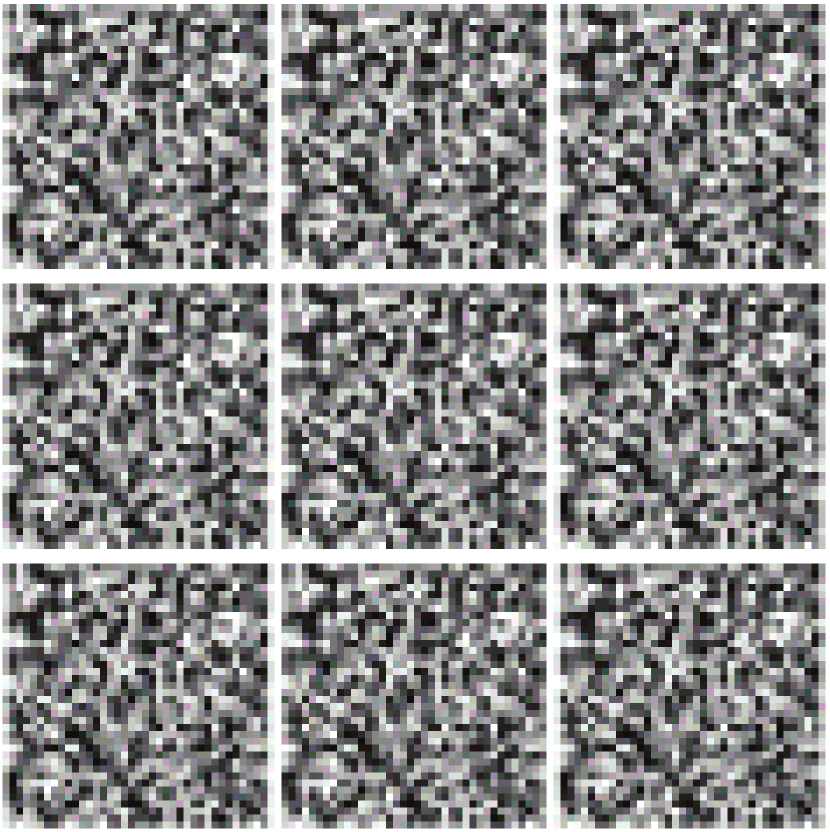}}

	\subfloat[DCCAE Normal Reconstruction]{
		\includegraphics[width=0.40\linewidth]{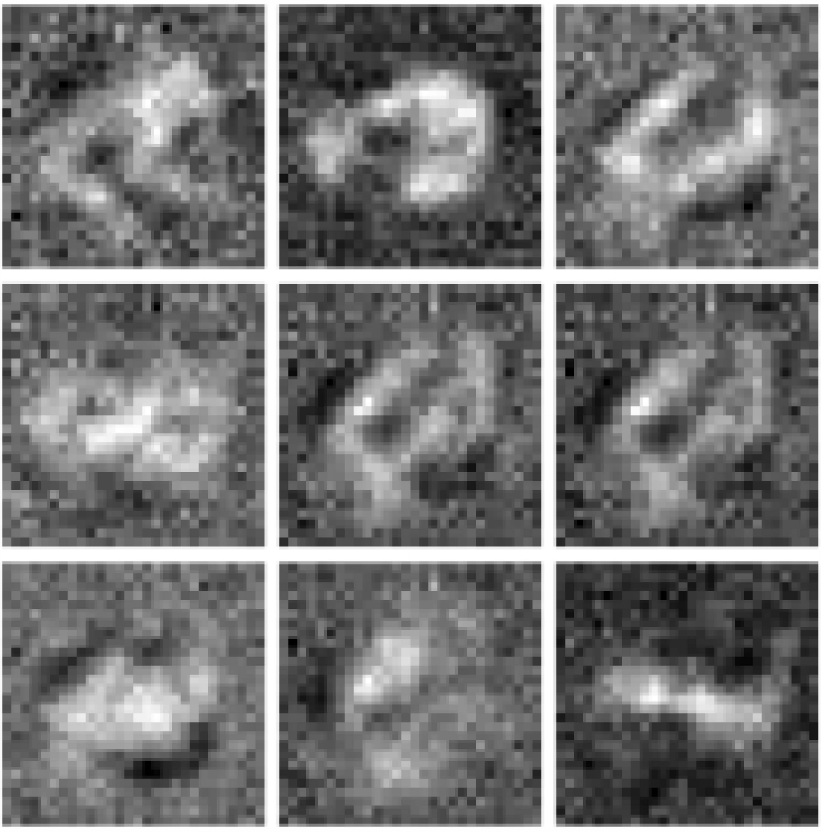}}
	\subfloat[DCCAE Noisy Reconstruction]{
		\includegraphics[width=0.40\linewidth]{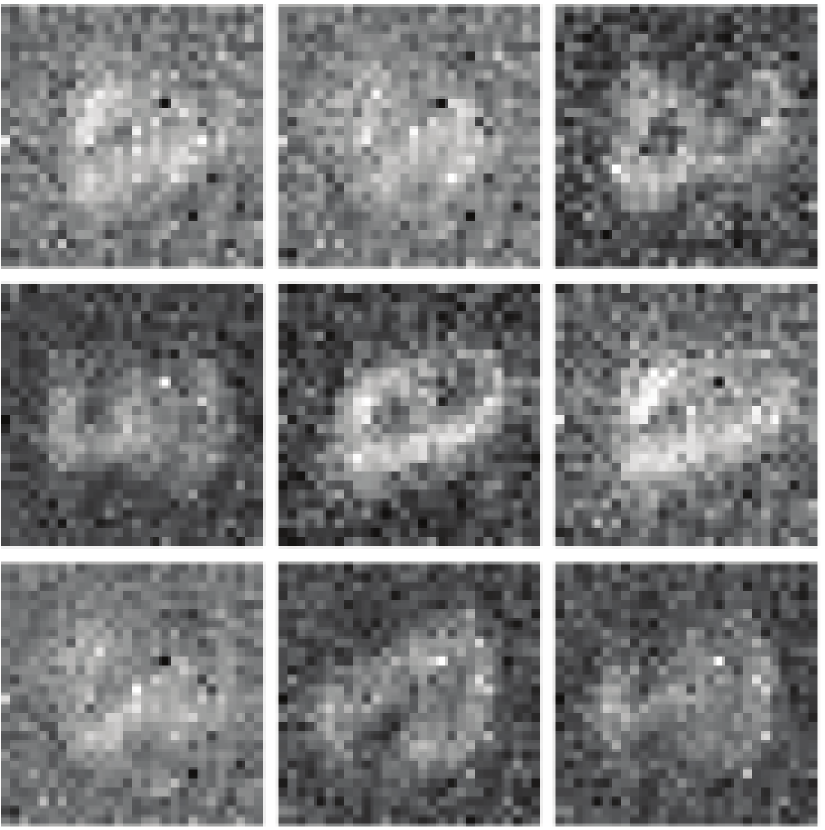}}
	
	\subfloat[${\text{CCDNN}_{\text{wRF}}}$ Normal Reconstruction]{
		\includegraphics[width=0.40\linewidth]{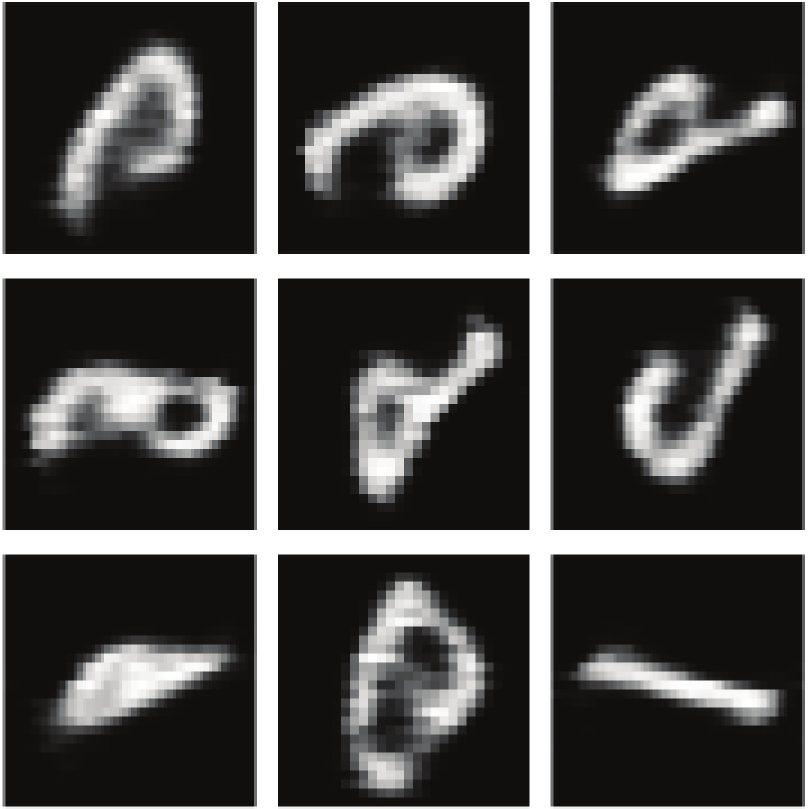}}
	\subfloat[${\text{CCDNN}_{\text{wRF}}}$ Noisy Reconstruction]{
		\includegraphics[width=0.40\linewidth]{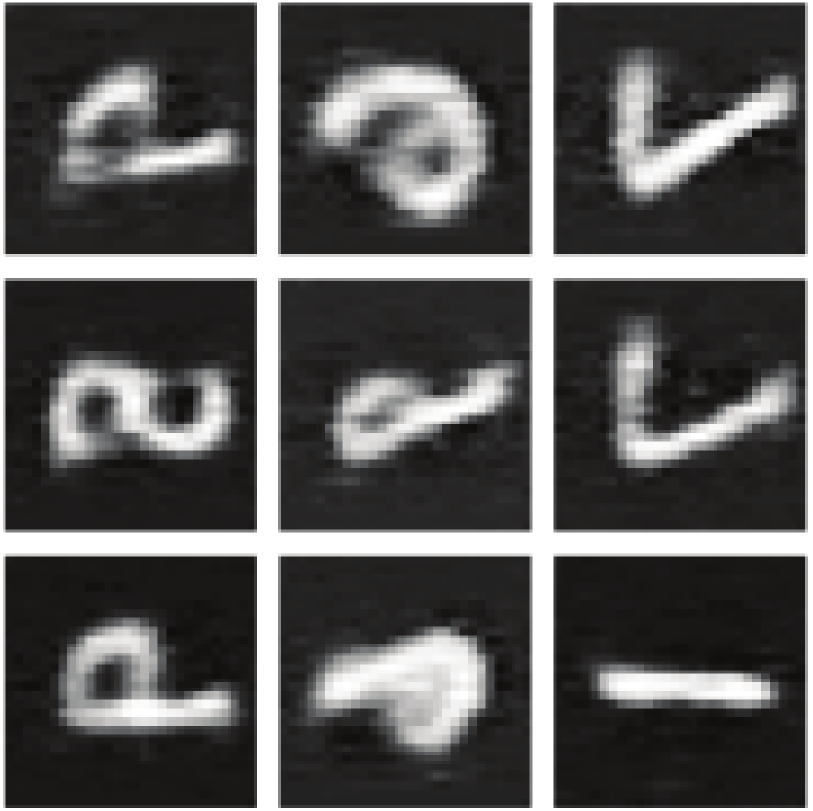}}
	
	\subfloat[CCDNN Normal Reconstruction]{
		\includegraphics[width=0.40\linewidth]{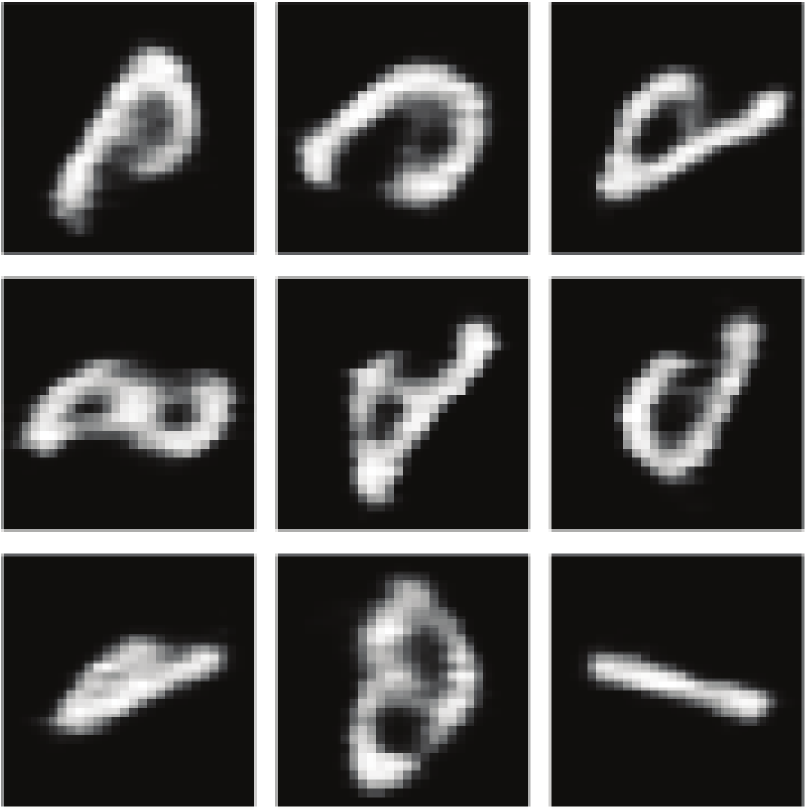}}
	\subfloat[CCDNN Noisy Reconstruction]{
		\includegraphics[width=0.40\linewidth]{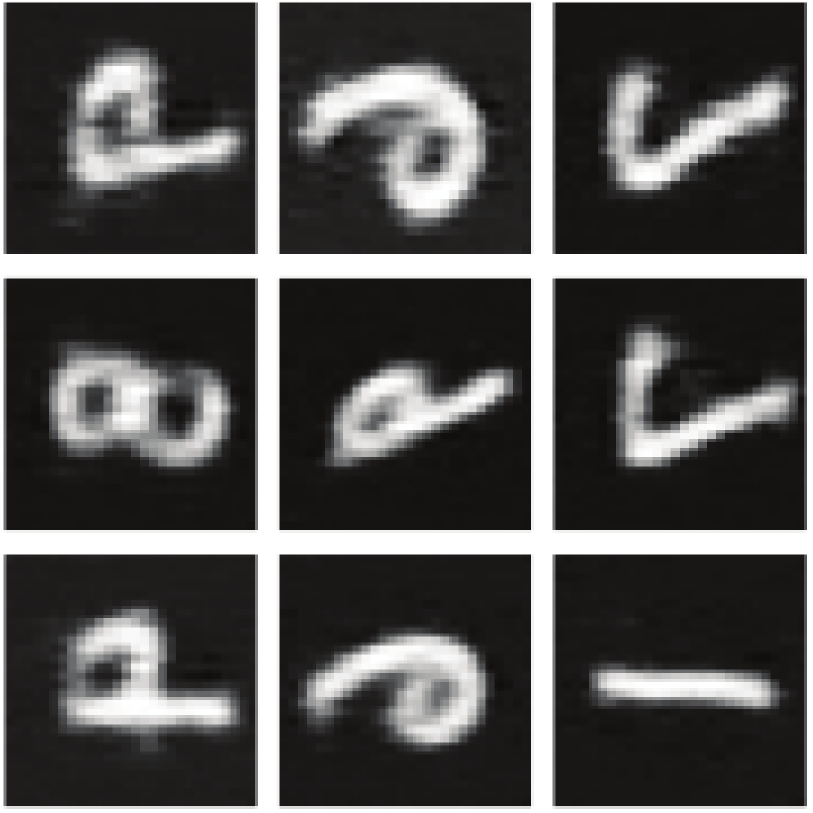}}
	
	\caption{\rmfamily Reconstruction results}
	\label{multi_images}
\end{figure}

The final reconstruction results are shown in Fig. \ref{multi_images} and Table \ref{t1}.

\begin{table}[h]
	\renewcommand{\arraystretch}{1.2}
	\caption{Performances of the Comparative Methods}
	\label{t1}
	\centering
	\resizebox{\columnwidth}{!}{
		\begin{tabular}{cccc}
			\hline\hline
			Method & MSE & MAE & Total Correlation \\ \hline
			DCCA & 0.53 $\pm$ 0.01 & 0.71 $\pm$ 0.01 & 6.55 $\pm$ 0.13 \\
			DCCAE & 0.19 $\pm$ 0.01 & 0.46 $\pm$ 0.01 & 6.60 $\pm$ 0.02 \\
			${\text{CCDNN}_{\text{wRF}}}$ & 0.12 $\pm$ 0.01 & 0.31 $\pm$ 0.01 & 2.20 $\pm$ 0.01 \\
			CCDNN & 0.10 $\pm$ 0.01 & 0.30 $\pm$ 0.00 & 2.25 $\pm$ 0.09 \\
			\hline\hline
		\end{tabular}
	}
\end{table}

As shown in Fig. \ref{multi_images}, CCDNN and ${\text{CCDNN}_{\text{wRF}}}$ outperform the two traditional deep CCA methods in terms of image reconstruction. This indicates that by using canonical correlation as a constraint rather than a single objective and adjusting the optimization goals for specific tasks, the network not only performs well in maximizing correlation but also better adapts to specific application needs.

As shown in Table \ref{t1}, the proposed CCDNN method has the smallest reconstruction error, with average MSE and MAE values of 0.10 and 0.30, respectively. Those values are 0.43 and 0.41 lower than the worst values of DCCA. Meanwhile, DCCAE shows improvement over DCCA due to the addition of an autoencoder regularization term that minimizes reconstruction error. The total correlation of the two methods under the canonical correlation guided learning framework is lower than that of the two deep CCA methods. In the image reconstruction task, a smaller total correlation indicates that the former two methods prioritize critical information over duplicating common information found across multiple views.

Compared to CCDNN with the redundancy filter, the reconstruction ability of ${\text{CCDNN}_{\text{wRF}}}$ has decreased, indicating that the model's capability to extract correlations is suppressed after the removal of correlation-induced redundancy. In denoising tasks, some redundant features may be part of the noise. Redundancy filter, by reducing the redundant information between views, can enhance the relevance between features and task objectives. Furthermore, by identifying and eliminating repetitive or similar features in multi-view data, this mechanism can reduce unnecessary computational burdens. This is beneficial for views with significant noise interference, helping enhance the model's denoising capabilities.

\subsection{Industrial Fault Diagnosis Task}
\begin{figure}[t]
	\begin{center}
		\includegraphics[width=8.4cm]{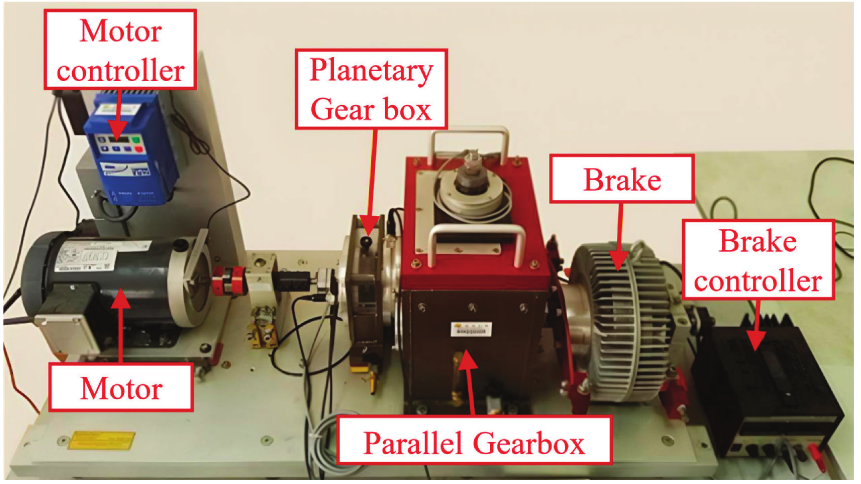}
		\caption{\rmfamily SEU Fault Diagnosis Platform}
		\label{SEU}
	\end{center}
\end{figure}

In this section, the bearing dataset from SEU is used for the industrial fault diagnosis task \cite{shao2018highly}. The test rig consists of several parts including a motor, motor controller, planetary gearbox, reduction gearbox, brake, and brake controller (to control the load), as shown in Fig. \ref{SEU}.

The fault operation conditions include two scenarios: speed 20Hz (1200rpm) - load 0V (0Nm) and speed 30Hz (1800rpm) - load 2V (7.32Nm). Data is collected from the Drive System Dynamics Simulator (DSDS), including motor vibration signals, planetary gearbox vibration signals in the x, y, and z directions, motor torque, and parallel gearbox vibration signals in the x, y, and z directions, with a sampling frequency of 5120Hz. We focus on four types of bearing faults and one normal operation condition: Ball fault (F1), Inner ring fault (F2), Outer ring fault (F3), Combination fault on both inner ring and outer ring (F4), and normal operation (Normal). The diagnostic capability is measured by Accuracy, defined as the percentage of correctly diagnosed samples out of the total number of samples.

To achieve the above goal, as shown in Fig. \ref{Task Network}, an CCDNN is constructed for the fault diagnosis task. The DNN part uses CNN. First, two sets of different sensor signals $\bm{x}_1$ and $\bm{x}_2$ are separately input into DNNs to obtain low-dimensional embeddings $\bm{z}_1$ and $\bm{z}_2$ mapped in the latent space; then both embeddings are input into the redundancy filter to obtain two residual signals $\bm{r}_1$ and $\bm{r}_2$; both of them are then concatenated to form a complete feature vector, which is fed into the Dense layer for further mapping; finally, an one-hot label $\bm{\hat {y}}_i$ is output through the Softmax layer for fault diagnosis. During training process, the parameters of the encoder and decoder are optimized by minimizing the cross entropy loss function:

\begin{align}
	\mathop {\arg \min }\limits_{({\bm{\theta} _1},{\bm{\theta} _2},{\bm{\theta} _3})}  - \sum\limits_{i} {{\bm{y}_i}\log {\bm{\hat {y}}_i}}  \label{e21}
\end{align}

\begin{figure}[t]
	\begin{center}
		\includegraphics[width=8.4cm]{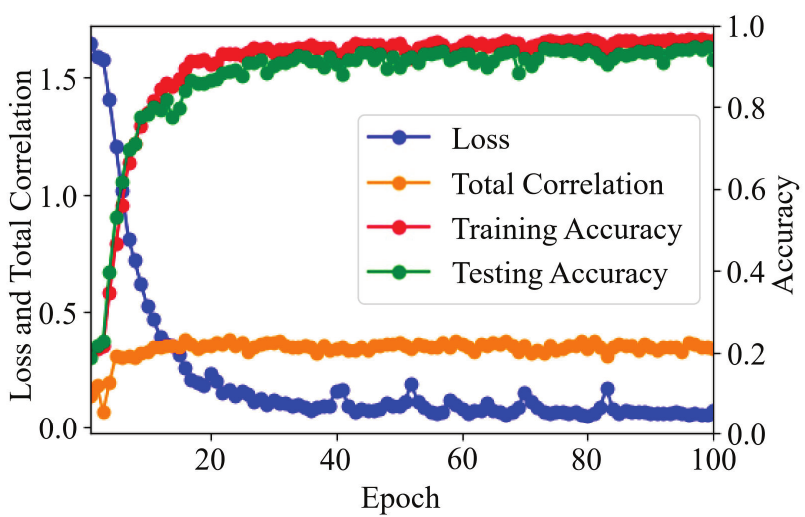}
		\caption{\rmfamily Training process of fault diagnosis}
		\label{FD training process}
	\end{center}
\end{figure}

Record the changes of multiple parameters within the first 100 epochs of CCDNN, as shown in Fig. \ref{FD training process}.

To verify the advantages of the proposed method in the fault diagnosis task, we compared the classical CNN method, the DCCAE method based on the deep CCA framework, and ${\text{CCDNN}_{\text{wRF}}}$. To explore the impact of parameter settings on the method, the performance of different methods under various training ratios, as well as the performance of the CCDNN method under three batch sizes, was recorded, as shown in Table \ref{FD Performance Metrics} and Table \ref{FD Batch Size}.

	\begin{table}[h]
		\renewcommand{\arraystretch}{1.2}
		\caption{Diagnostic Effects of Different Methods}
		\label{FD Performance Metrics}
		\centering
		\resizebox{\columnwidth}{!}{
			\begin{tabular}{cccc}
				\hline\hline
				Methods & Batch Size & Training Ratio & Accuracy (\%) \\ \hline
				\multirow{5}{*}{CNN} & \multirow{5}{*}{256} & 0.4 & 82.76 $\pm$ 0.19 \\
				& & 0.5 & 86.62 $\pm$ 0.14 \\
				& & 0.6 & 88.34 $\pm$ 0.01 \\
				& & 0.7 & 89.39 $\pm$ 0.01 \\
				& & 0.8 & 94.79 $\pm$ 0.03 \\ \hline
				\multirow{5}{*}{DCCA} & \multirow{5}{*}{256} & 0.4 & 38.80 $\pm$ 0.40 \\
				& & 0.5 & 39.47 $\pm$ 0.30 \\
				& & 0.6 & 37.66 $\pm$ 0.66 \\
				& & 0.7 & 35.51 $\pm$ 1.02 \\
				& & 0.8 & 38.54 $\pm$ 2.90 \\ \hline
				\multirow{5}{*}{{$\text{CCDNN}_\text{wRF}$}} & \multirow{5}{*}{256} & 0.4 & \textbf{91.53 $\pm$ 0.01} \\
				& & 0.5 & 93.08 $\pm$ 0.12 \\
				& & 0.6 & 92.98 $\pm$ 0.05 \\
				& & 0.7 & 96.19 $\pm$ 0.01 \\
				& & 0.8 & 97.57 $\pm$ 0.01 \\ \hline
				\multirow{5}{*}{CCDNN} & \multirow{5}{*}{256} & 0.4 & 91.19 $\pm$ 0.05 \\
				& & 0.5 & \textbf{94.02 $\pm$ 0.07} \\
				& & 0.6 & \textbf{95.04 $\pm$ 0.01} \\
				& & 0.7 & \textbf{97.26 $\pm$ 0.01} \\
				& & 0.8 & \textbf{97.79 $\pm$ 0.01} \\
				\hline\hline
			\end{tabular}
		}
	\end{table}

\begin{table}[h]
	\renewcommand{\arraystretch}{1.2}
	\caption{Comparison of Effects under Different Batch Sizes}
	\label{FD Batch Size}
	\centering
	\resizebox{\columnwidth}{!}{
		\begin{tabular}{cccc}
			\hline\hline
			Methods & Training Ratio & Batch Size & Accuracy (\%) \\ \hline
			\multirow{3}{*}{CCDNN} & \multirow{3}{*}{0.7} & 64 & 98.92 $\pm$ 0.33 \\
		&	& 128 & 95.17 $\pm$ 0.04 \\
		&	& 256 & 97.26 $\pm$ 0.01 \\
			\hline\hline
		\end{tabular}
	}
\end{table}
As shown in Table \ref{FD Performance Metrics}, the two CCDNN methods have higher diagnostic accuracy and smaller variance, outperforming the CNN and DCCAE methods. Smaller variance indicates that the CCDNN model exhibits stronger stability in distinguishing different fault types in complex real-world applications. The DCCAE model performs poorly in this experiment, possibly due to its limited ability to handle complex data, especially without enhancing its feature processing capability through deep learning architecture. Under a smaller training ratio (0.4), ${\text{CCDNN}_{\text{wRF}}}$ performs better, while as the training ratio increases, the complete CCDNN performs better. Overall, the difference between them is slight. As shown in Table \ref{FD Batch Size}, the diagnostic accuracy is highest when the batch size is 64, but as it increases to 128 and 256, the accuracy first increases and then decreases. From this point, we need to choose the appropriate batch size according to different tasks to optimize the model's performance.

\begin{figure}[t]
	\centering
	\subfloat[DCCAE t-SNE]{
		\includegraphics[width=0.45\linewidth]{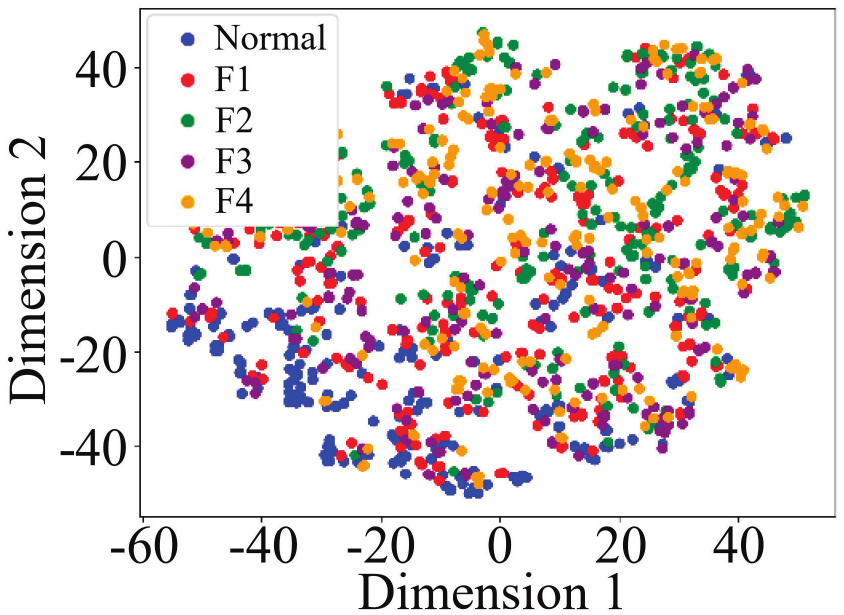}}
	\subfloat[CNN t-SNE]{
		\includegraphics[width=0.45\linewidth]{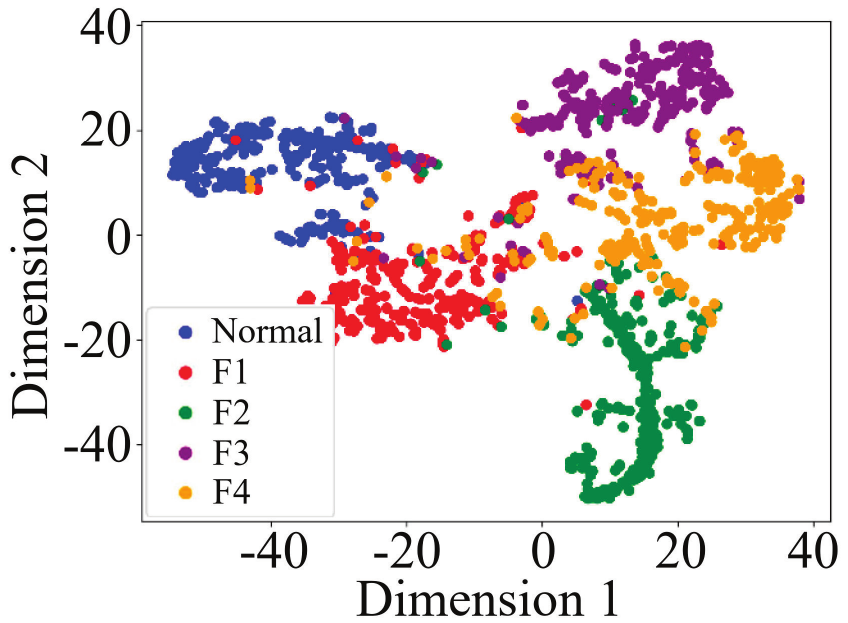}}
	
	\subfloat[${\text{CCDNN}_{\text{wRF}}}$ t-SNE]{
		\includegraphics[width=0.45\linewidth]{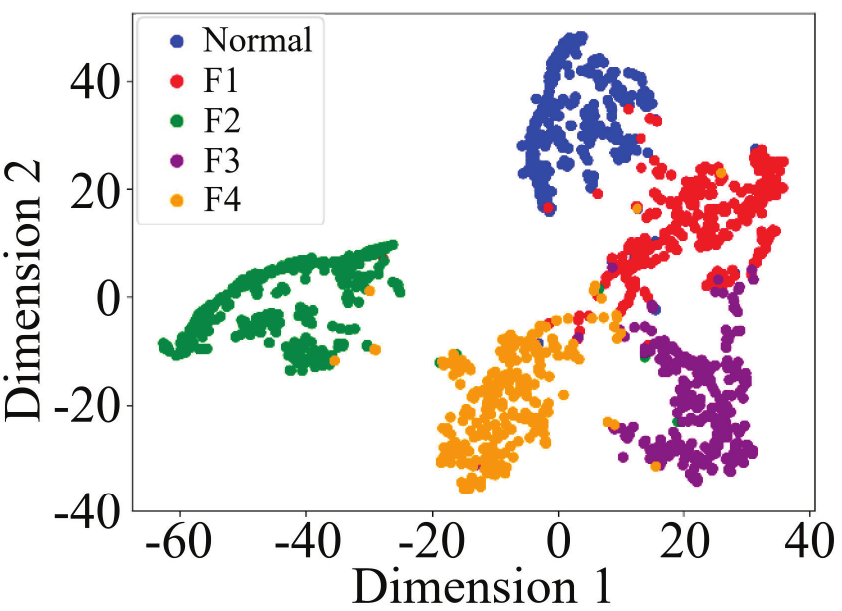}}
	\subfloat[CCDNN t-SNE]{
		\includegraphics[width=0.45\linewidth]{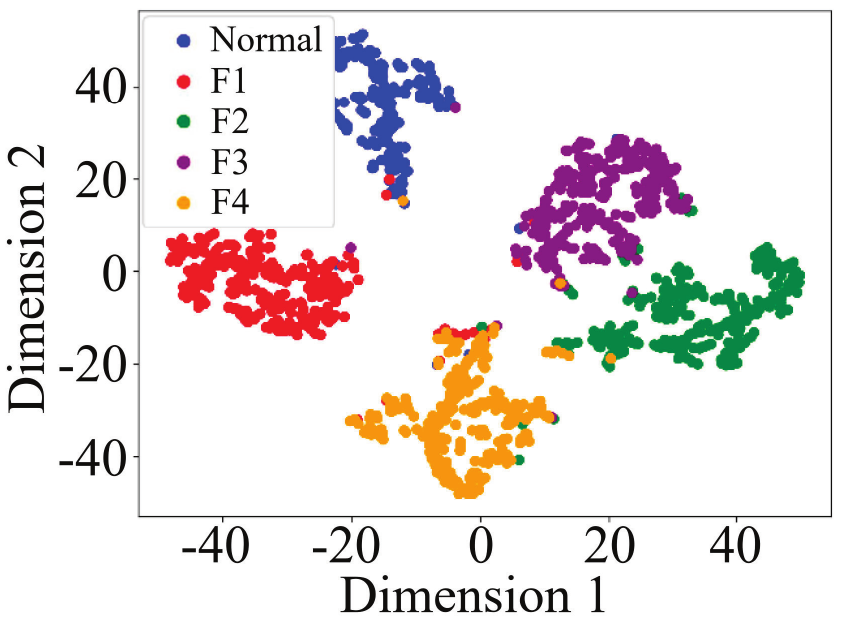}}
	
	\caption{\rmfamily Visualization via t-SNE}
	\label{F_TSNE}
\end{figure}

To further analyze the impact of the redundancy filter on CCDNN in the fault diagnosis task, we also qualitatively studied the features by embedding the projected features in 2D using t-SNE, and the visualization results are shown in Fig. \ref{F_TSNE}. Overall, the class separation in the visualization qualitatively corresponds to the diagnostic performance in Table \ref{FD Performance Metrics}. Observing the t-SNE plot, it can be seen that the CCDNN with the redundancy filter achieves better class separation in the feature space, with clearer boundaries. Although the diagnostic accuracy of both methods is close under the current experimental settings, the CCDNN with the redundancy filter shows more distinct feature mapping points for different fault types in the feature space, indicating greater differentiation between categories. Additionally, the boundary between normal and fault samples is clear, reducing the rate of false alarms.

\begin{figure}[!t]
		\begin{center}
			\includegraphics[width=8.4cm]{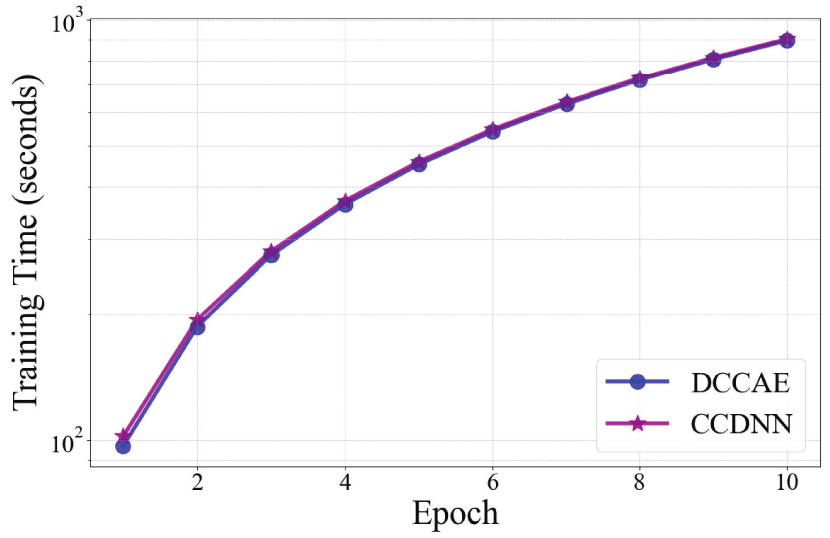}
			\caption{\rmfamily Comparison of training speed in terms of time-consuming}
			\label{FD training time}
		\end{center}
	\end{figure}

Furthermore, to assess the training speed of CCDNN, a numerical experiment is conducted with a comparison of DCCAE. The experimental setup consisted of an Intel Xeon Silver 4116 processor and eight NVIDIA GeForce RTX 2080 Ti graphics cards. Specifically, under the condition of batch size=256, the cumulative time of 10 epochs are recorded as shown in Fig. \ref{FD training time}. As can be seen from that, CCDNN has almost the same training time with DCCAE. This demonstrate that the new configuration does not bring additional computational burden, which preserves the correlated representation learning ability and focuses more on the engineering tasks.

\subsection{Remaining Useful Life Task}

In this subsection, the NASA Ames Prognostics Data Repository dataset is used for case of remaining useful life (RUL) \cite{saxena2008turbofan}. This dataset includes time-series readings from 24 sensors across 100 turbofan engines from the start of use to the end of their useful life. There are three operational condition variables as settings and 21 sensor readings as monitoring variables. Before building the prediction model, feature selection is required based on the correlation between each sensor reading and the engine degradation process. Historical data is divided into time-series of a certain length to create training and testing sets for different models. The distribution of monitoring values changes under different conditions, so the data is divided into six subsets according to working states, and the sensor readings in each subset are normalized using the Z-Score normalization method. After normalization, some sensor data show obvious trends, so the next step is to select sensors with obvious trends. The feature selection method used in this paper is the least squares method, which fits a straight line to the time series of each sensor and selects the top eight sensors with the largest absolute slopes of the lines. The selected sensors from the dataset are [2, 3, 4, 7, 8, 11, 12, 15], and the other sensor data are discarded. For each prediction, the historical data amount is set to 50 cycles, each cycle containing readings from the aforementioned eight sensors, and the original data is converted into time-series through a sliding window.

As shown in Fig. \ref{Task Network}, an CCDNN is constructed for the RUL task. Given that RUL is a time-series-related task, the experiment not only uses CNN but also replaces CNN with GRU in the DNN module for comparison. The DNN module extracts features $\bm{z}_1$ and $\bm{z}_2$, which then pass through the CCA layer and the redundancy filter to obtain residual signals $\bm{r}_1$ and $\bm{r}_2$. After concatenation, they are input into the dense layer. Unlike the classification task of fault diagnosis, RUL is a regression task, and the output of the dense layer is the predicted vector ${{{\bm{\hat{y}}}_i}}$. The CCDNN architecture allows us to adapt to different tasks by modifying the output layer while keeping the core parts unchanged. During training process, the parameters are optimized by minimizing the mean squared error loss function.
\begin{align}
	\mathop {\arg \min }\limits_{({\bm{\theta} _1},{\bm{\theta} _2},{\bm{\theta} _3})} \frac{1}{n} \sum\limits_i^n {{{{|| {{\bm{y}_i} - {{\bm{\hat{y}}}_i}} ||}^2}} }  \label{e23}
\end{align}

\begin{figure}[t]
	\begin{center}
		\includegraphics[width=8.4cm]{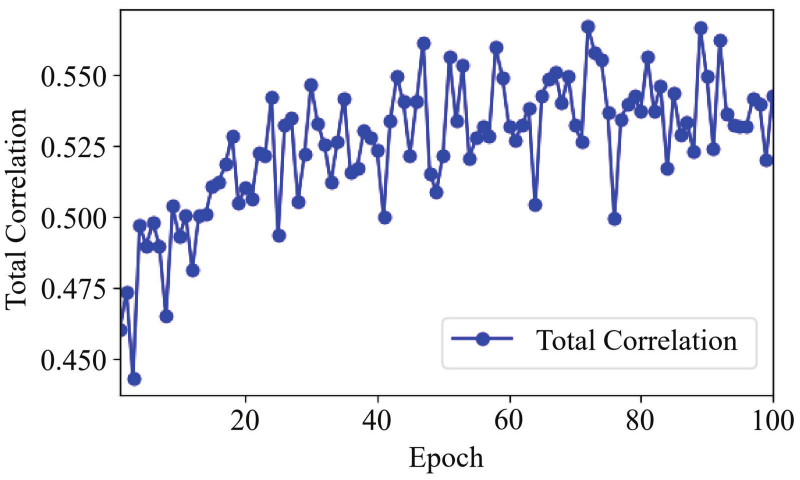}
		\caption{\rmfamily Total correlation during the training process}
		\label{RUL training proceess}
	\end{center}
\end{figure}

The total correlation of CCDNN for the first 100 epochs was recorded, and the changes during the training process are shown in Fig. \ref{RUL training proceess}.

\begin{figure}[t]
	\begin{center}
		\includegraphics[width=8.4cm]{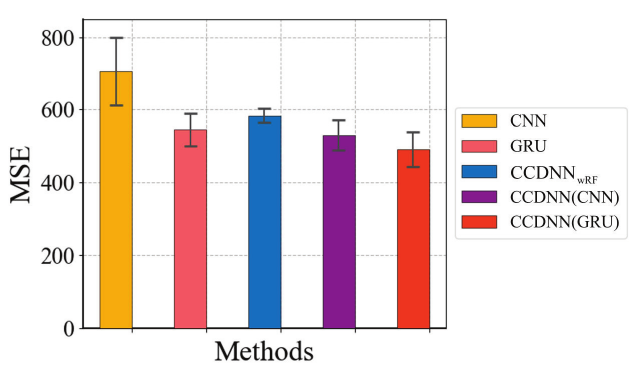}
		\caption{\rmfamily Comparison result of RUL in term of MSE}
		\label{RUL_result_MSE}
	\end{center}
\end{figure}

\begin{figure}[h!tb]
	\begin{center}
		\includegraphics[width=8.4cm]{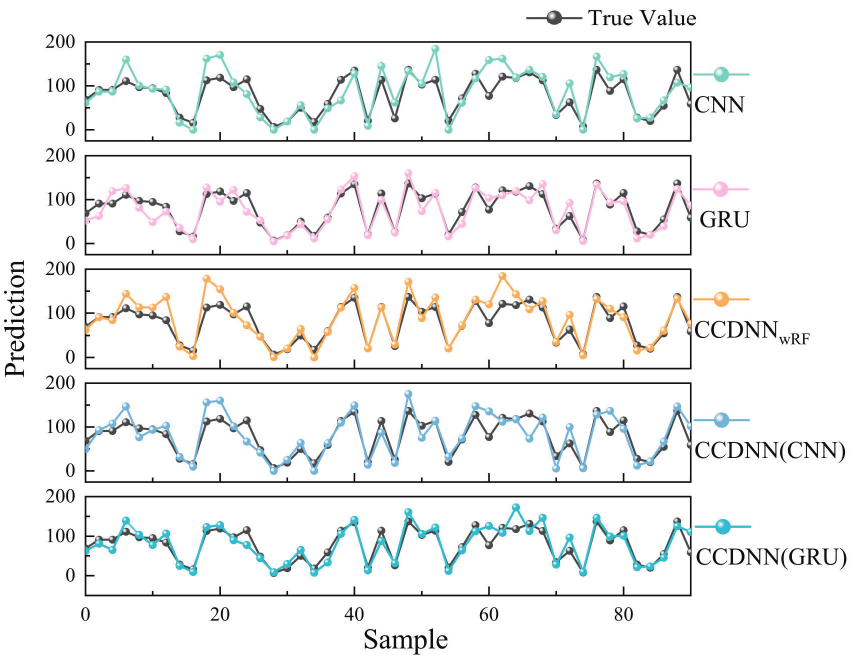}
		\caption{\rmfamily RUL prediction curves}
		\label{RUL result2}
	\end{center}
\end{figure}

To verify the advantages of the proposed method in the RUL task, we compared two classical RUL methods: CNN and GRU, as well as CCDNN without the redundancy filter. The metric for evaluating the reconstruction performance of those methods is MSE, and the final results are shown in Fig. \ref{RUL_result_MSE} and Fig. \ref{RUL result2}.

It can be seen that CCDNN (GRU) has the lowest MSE, being the closest to the true curve in predicting trend changes, achieving the best prediction effect. Overall, the GRU method is more suitable for the RUL task compared to the CNN method. This is because CNN tends to extract static, spatial features from the input, whereas GRU is better at extracting temporal features. GRU can better analyze the dynamic characteristics and long-term dependencies in time-series data, which exactly meets the requirements of the RUL task.

After introducing the canonical correlation guided learning framework, the MSE values of CNN and GRU decreased by 175.74 and 53.93, respectively, which demonstrates the importance of the CCA constraint in the proposed framework. The prediction accuracy of CCDNN without the redundancy filter significantly decreased, indicating that its ability to extract correlated features is weakened. The overlapping information recorded by sensors tends to concentrate in untreated features, leading to high redundancy. The redundancy filter can reduce the model's burden of processing irrelevant features without adding computational load, enhancing the model's sensitivity to key features.

\section{Conclusions}
In this paper, a canonical correlation guided deep neural network is proposed by merging multivariate analysis and machine learning. Unlike the standard CCA, kernel CCA and deep CCA methods, the optimization formulation of the proposed method is not restricted to maximize correlation, instead canonical correlation is used as a constraint, which preserves the correlated representation learning ability and focuses more on the optimization formulation. Then, to reduce the redundancy induced by correlation, a redundancy filter is designed and it has zero trainable parameters. The experimental results on MINST dataset show that the proposed method has better construction performance by comparing with DCCA and DCCAE. Furthermore, the application of the proposed method to fault diagnosis and remaining useful life cases shows that it has better performance in both tasks by comparing with existing methods.

CCDNN can learn flexible nonlinear representations via DNNs, hence, how to select appropriate DNNs for specific engineering task is worth studying. Moreover, both views of data are also flexible, which enables CCDNN to deal with multi-source heterogeneous data structure with different industrial applications, for instance, an engineering task of fault diagnosis, in which a view is given by images and the other view is given by time-series.

\section*{Acknowledgement}
The authors would like to sincerely thank Mr. Ketian Liang, Prof. Yun Wang and Prof. Xiaojun Zhou for invaluable discussion, reviewing and editing the paper.

\section*{Appendix: Extension of CCDNN}
	
\begin{figure*}[t]
		\begin{center}
			\includegraphics[width=15.4cm]{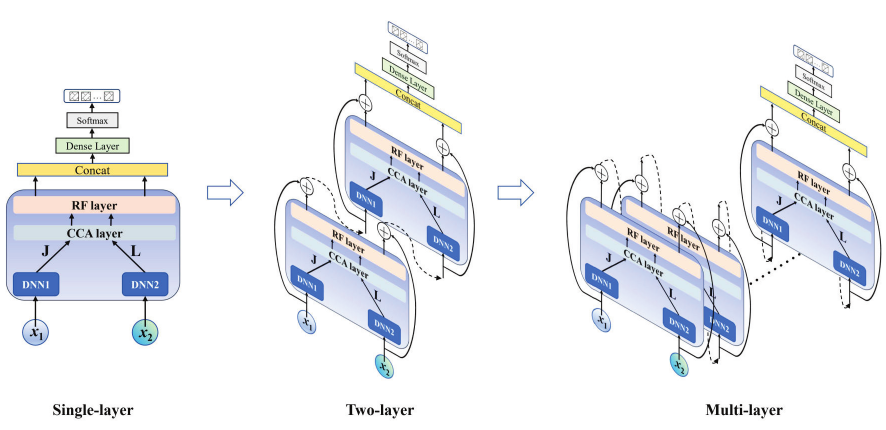}
			\caption{\rmfamily Extension of CCDNN structure}
			\label{Task Network_res}
		\end{center}
	\end{figure*}

The standard CCDNN network mainly includes three core layers: the DNN layer for processing the two input views, the CCA layer for maximizing the correlation between outs of two DNNs, as well as the RF layer for removing the redundancy induced by correlation, with the output features $\bm{r}_1$ and $\bm{r}_2$ used for subsequent tasks, as shown in the left subfigure of Fig. \ref{Task Network_res}. For better understanding, we name this standard CCDNN as `single-layer' CCDNN .
	
Building on the single-layer CCDNN, we attempt a multi-layer network architecture. Specifically, the core three-layer network of the proposed CCDNN is encapsulated into a basic network module, including the DNN layer, the CCA layer, and the RF layer. When both inputs, $\bm{x}_1$ and $\bm{x}_2$, are feed into this network module, the features extracted by the RF layer are output. Unlike the traditional approach of concatenating and then outputting to the dense layer, we feed these features back into the network module as new features to be learned. After multi-layer learning and feature extraction, the output is used for subsequent tasks. Additionally, due to the increased network complexity and the RF output being the key features after redundancy removal, to prevent issues like gradient vanishing during training, we introduce a residual structure \cite{Kaiming_ResNet}. Specifically, the output of one layer is concatenated with its input as the input data for the next layer. The equation of residual concatenation is as follows,

	\begin{alignat}{2}
		& \bm{x}_{1i} = (\mathbf{J}_i^{\text{T}}f_1(\bm{x}_{1(i - 1)}) - \mathbf{\Sigma}_i \mathbf{L}_i^{\text{T}}f_2(\bm{x}_{2(i - 1)})) \oplus \bm{x}_{1(i - 1)} \nonumber \\
		& \bm{x}_{2i} = (\mathbf{J}_i^{\text{T}}f_2(\bm{x}_{2(i - 1)}) - \mathbf{\Sigma}_i^{\text{T}}\mathbf{J}_i^{\text{T}}f_1(\bm{x}_{1(i - 1)})) \oplus \bm{x}_{2(i - 1)}
	\end{alignat}
where $\bm{x}_{1i}$ and $\bm{x}_{2i}$ represent the inputs of the $i$-th layer of DNN1 and DNN2, respectively, and $\bm{x}_{1(i-1)}$ and $\bm{x}_{2(i-1)}$ represent the inputs of the $(i-1)$-th layer. $\mathbf{J}_i$, $\mathbf{\Sigma}_i$, and $\mathbf{L}_i$ are the outputs of the two views extracted by CCA in the $i$-th layer, and $\oplus$ represents concatenation. The appropriate concatenation dimension can be selected based on different features.
	
Taking the classification task as an example, the structure from a single-layer to a two-layer and then to a multi-layer network is shown in Fig. \ref{Task Network_res}.

	To validate the effectiveness of the proposed method, we conducted further experiments based on the previous SEG fault diagnosis dataset. Considering the training speed of the network, we tested a two-layer network, keeping all other hyperparameters consistent with the single-layer network described earlier. The results are shown in Table \ref{t11}.
	
	\begin{table}[h]
		\renewcommand{\arraystretch}{1.2}
		\caption{Performances of the Comparative Methods}
		\label{t11}
		\centering
		\resizebox{\columnwidth}{!}{
			\begin{tabular}{cccc}
				\hline\hline
				Method & Batch size & Training Ratio & Accuracy (\%) \\ \hline
				CNN & 256  & 0.7 & 89.39 $\pm$ 0.01 \\
				DCCA & 256  & 0.7 & 35.51 $\pm$ 1.02 \\
				${\text{CCDNN}_{\text{wRF}}}$ & 256  & 0.7 & 96.19 $\pm$ 0.01 \\
				CCDNN & 256  & 0.7 & 97.26 $\pm$ 0.01 \\
				\textbf{${\text{CCDNN}_{\text{d}}}$} & 256  & 0.7 & \textbf{98.46 $\pm$ 0.01} \\
				\hline\hline
			\end{tabular}
		}
	\end{table}

It can be seen that the two-layer CCDNN method is superior to the single-layer CCDNN method in terms of diagnostic accuracy, indicating that it has better feature extraction capabilities. Additionally, the introduction of the residual concatenation method allows the input features of the previous layer to combine with the redundant features processed by the RF layer. This combination alleviate information loss and the common gradient vanishing problem in deep neural networks, making the training process more stable.
	
In this paper, we only indicate the usage of the two-layer CCDNN, as shown in the middle subfigure of Fig. \ref{Task Network_res}. In the future, multi-layer CCDNN will be used according to the demand of tasks and the complexity of the available datasets.

\bibliographystyle{Bibliography/IEEEtranTIE}
\bibliography{Bibliography/IEEEabrv,Bibliography/centralBib}\

\end{document}